# Evaluating Deep Neural Network Ensembles by Majority Voting cum Meta-Learning scheme

Anmol Jain, Aishwary Kumar, Seba Susan[0000-0002-6709-6591]*

Department of Information Technology,
Delhi Technological University,
Bawana Road, Delhi, India-110042
seba_406@yahoo.in

**Abstract.** Deep Neural Networks (DNNs) are prone to overfitting and hence have high variance. Overfitted networks do not perform well for a new data instance. So instead of using a single DNN as classifier we propose an ensemble of seven independent DNN learners by varying only the input to these DNNs keeping their architecture and intrinsic properties same. To induce variety in the training input, for each of the seven DNNs, one-seventh of the data is deleted and replenished by bootstrap sampling from the remaining samples. We have proposed a novel technique for combining the prediction of the DNN learners in the ensemble. Our method is called *pre-filtering by majority voting coupled with stacked meta-learner* which performs a two-step confidence check for the predictions before assigning the final class labels. All the algorithms in this paper have been tested on five benchmark datasets namely, Human Activity Recognition (HAR), Gas sensor array drift, Isolet, Spambase and Internet advertisements. Our ensemble approach achieves higher accuracy than a single DNN and the average individual accuracies of DNNs in the ensemble, as well as the baseline approaches of plurality voting and meta-learning.

## 1    Introduction

Deep Neural Network (DNN) has multiple hidden layers and each hidden layer has hundreds or thousands of activation units present in it [1]. When we use DNN as the classifier, issues of computational expense and overfitting of data may crop up. DNNs usually exhibit high variance for small real-word datasets. Because of high variance, when a novel data instance is fed, the model does not perform well. So instead of a single DNN, we propose to use an ensemble of multiple DNNs [2], each of them making independent errors, and we can combine their predictions in some manner to get a better model. Using the DNN ensemble instead of a single DNN also minimizes, to some extent, the problem of convergence to local minima due to gradient descent optimization [3, 4]. Alternative solutions to usage of ensembles for inducing variety in learning include the global optimization of network weights using evolutionary algorithms such as Particle Swarm Optimization [5].

Suppose we have *n* number of independent DNN learners and let $m_j$ be the output of the $j^{th}$ learner. The combined variance of the net ensemble output y can be written as

$$\text{Var}(y) = \text{Var}\left(\sum_j \frac{1}{n} * m_j\right) \qquad (1)$$

which can be rewritten as in (2) and (3).

$$\text{Var}(y) = \frac{1}{n^2} * n * \text{Var}(m_j) \qquad (2)$$

$$\text{Var}(y) = \frac{1}{n} * \text{Var}(m_j) \qquad (3)$$

i.e. the variance of an ensemble gets reduced by a factor of *n* assuming that the *n* learners are uncorrelated. The entire concept of getting a good ensemble is to get independent learners instead of individual good learners. We can get independent learners in an ensemble by sampling the training set with replacement, also called bootstrap aggregation or bagging [6]. We create a subsample of size *s* from the initial dataset of size *n*. Sampling is done in such a way that the subsample of size *s* is identically and independently distributed (IID) and can be considered as representative for the whole sample. It is also possible to have different training subsets by using feature selection method, selecting only a subset of attributes to train each learner [7], resulting in diverse and independent learners. In some cases, where the dataset is small, some random noise could be introduced (such as gaussian noise) in the training data, to get independent leaners, that reduces the generalization error [8]. Adding noise to the training data contributes to a regularization factor that reduces the variance. It is also possible to have DNNs with different architectures and varying hyper-parameters such as different number of hidden layers and activation units in each layer [9], batch size, number of epochs, activation functions etc. In our paper, we investigate DNN ensembles with induced variety in training input, and explore various techniques of combining the DNN outputs in an effective manner. The organization of this paper is as follows. Basic ensemble concepts are revisited in section 2, the proposed DNN ensemble and learning methodology is presented in section 3, the experimentation and the results are discussed in section 4, and the final conclusions are given in section 5.

## 2     Combining the results of independent learners

Once we have a set of independent learners, the next step is to combine their predictions to get better predictions than the individual trained models. There are various approaches for fusing the outcomes of the *n* learners in an ensemble. We review several of the popular decision-fusion techniques next.

(i) *Model averaging or unweighted voting.* In this approach, the predicted probabilities of a class, from all the independent learners, are summed up and the final prediction is made by taking the maximum of all the probabilistic sums of predicted classes [10]. This works fine when we have a good estimation of probabilities. However, this method does not incorporate inter-classifier and intra-classifier biases.

(ii) *Weighted model averaging or weighted voting.* When combining the predictions from the learners, there are some learners that are more significant as compared to the other learners. In that case, we allot weights to predictions corresponding to the significance-level of each independent learner [11]. The optimal weights can be assigned by using gradient descent optimization procedure or using grid search. Simply, we can also assign weights proportional to accuracies of individual DNNs (Eq. (4)) or inversely proportion to the variances contributed by them (Eq. (5)).

$$weight \propto accuracy \tag{4}$$

$$weight \propto \frac{1}{variance} \tag{5}$$

However, some results also suggest that choosing optimized weights results in loss of generalization because of overfitting.

(iii) *Plurality voting.* In this approach each learner makes a prediction for a particular class label and the final candidate is chosen which has got the maximum votes [12]. Suppose we have three class labels $c_1$, $c_2$ and $c_3$ and out of *n* learners $n_1$ learners predict $c_1$, $n_2$ predict $c_2$ and $n_3$ predict $c_3$. So, the final prediction is given as argmax ($n_1$, $n_2$, $n_3$).

(iv) *Majority voting.* Slightly distinct from plurality voting due the incorporation of a threshold for the maximum votes obtained, here, every learner makes individual predictions and the candidate for final prediction is the one which gets more than half of the total votes [13]. If none of the class labels qualifies the criteria then that test instance is considered as an error.

(v) *Meta-learner classifier.* The outcomes from all the base learners in the ensemble are treated as level 1 predictions; these predictions are fed as input to a meta-learner classifier [14] (also called level 2 learner). The outcomes from level 2 are treated as the final outcomes for a particular test instance. A most common instance in machine learning is using a meta-learner classifier for learning the hidden state activations (feature-vectors) of independent learners, after fusing the features by concatenation [15]. In ensemble learning, we fuse the predicted outputs of independent learners rather than their hidden state activations. Meta-learners include Neural Networks or fully connected dense layers [16], AdaBoost [17] and XGBoost [18].

## 3 Proposed ensemble approach

Our homogenous ensemble comprises of seven identical DNNs having two hidden layers with 1200 and 800 activations units, respectively. We followed a novel approach to diversify DNN learning by deleting 1/7$^{th}$ of the training data and replenishing the lost samples by randomly replicating the remaining samples, a concept known as bootstrap sampling [19]. The exercise is repeated for all seven DNNs in the ensemble (each time a different 1/7$^{th}$ segment is deleted) to ensure variety in the training input, as shown in Fig. 1 (a). Different fusion strategies for combining the DNN outputs were tried in order to reduce the overall variance of the ensemble and to get better results in the final prediction. These are described next in the order from (i) to (iii).

(i) In the first experiment, we implemented plurality voting. We took the maximum of the probabilities predicted by a DNN for each class label and assigned the class label corresponding to that probability value to the test data. Output class labels from each of the DNNs were polled. We store the frequency of each class label as the result of polling. Finally, the test instance is allotted the class label with the maximum frequency.

$$Class = \mathrm{argmax}(\mathrm{fre}[y_k]) \tag{6}$$

where the array fre[$y_k$] stores the count of class label $y_k$ and *Class* is the final class label predicted by our ensemble for the test instance.

(ii) In the second experiment, we used a meta-learner XGBoost for learning the DNN predictions and predicting the ensemble output. The resampled training set was given as input to the DNNs in the ensemble and their corresponding outputs were combined (level 1 prediction) as a feature-vector that was then used to train the meta-learner classifier XGBoost. Then for the test set, we first got level 1 prediction and then for the final ensemble output we passed it to the meta-learner.

(iii) The third experiment involves the proposed *pre-filtering by majority voting coupled with a stacked meta-learner* approach. When using the polling-based method to track count of each class label and determining the class label having maximum count, there could be cases when multiple class labels have the same frequency and also get the same share of maximum votes. In those cases, the polling-based method usually selects any one of those class labels as the final prediction. To make decisions in such cases, we first perform filtering of level 1 predictions using majority voting and then use the stacked meta-learner for the cases with no clear majority. The predictions having count of class label greater than or equal to *n*-1, where *n* is the ensemble size, were left as it is and the rest were filtered out and fed as input to the meta-learner. Addition of a filtering stage improves the performance of the meta-learner which is now trained on difficult cases only. If at least *n*-1 learners out of *n* predicted the same class for a given instance, we were highly confident of the prediction. The remaining filtered instances were passed to a meta-learner for further surety, which is a better approach for resolving conflicts of maximum votes. The training process for the meta-learner is shown in Fig. 1 (b). The test sample is presented to all the DNNs in the ensemble, and the DNN outputs are subject to majority voting. If the maximum number of votes is less than *n*-1, the DNN predictions are given as an input feature vector to the trained meta-learner for predicting the class label. The process flow for the test instance is shown in Fig. 2. We also experimented by varying the number of DNNs in our ensemble from one to eight with step-size of one. Accuracies were recorded for each ensemble and the optimal ensemble size was determined. The graphs in Fig. 3 indicate that a choice of seven DNNs is optimal for our experiments.

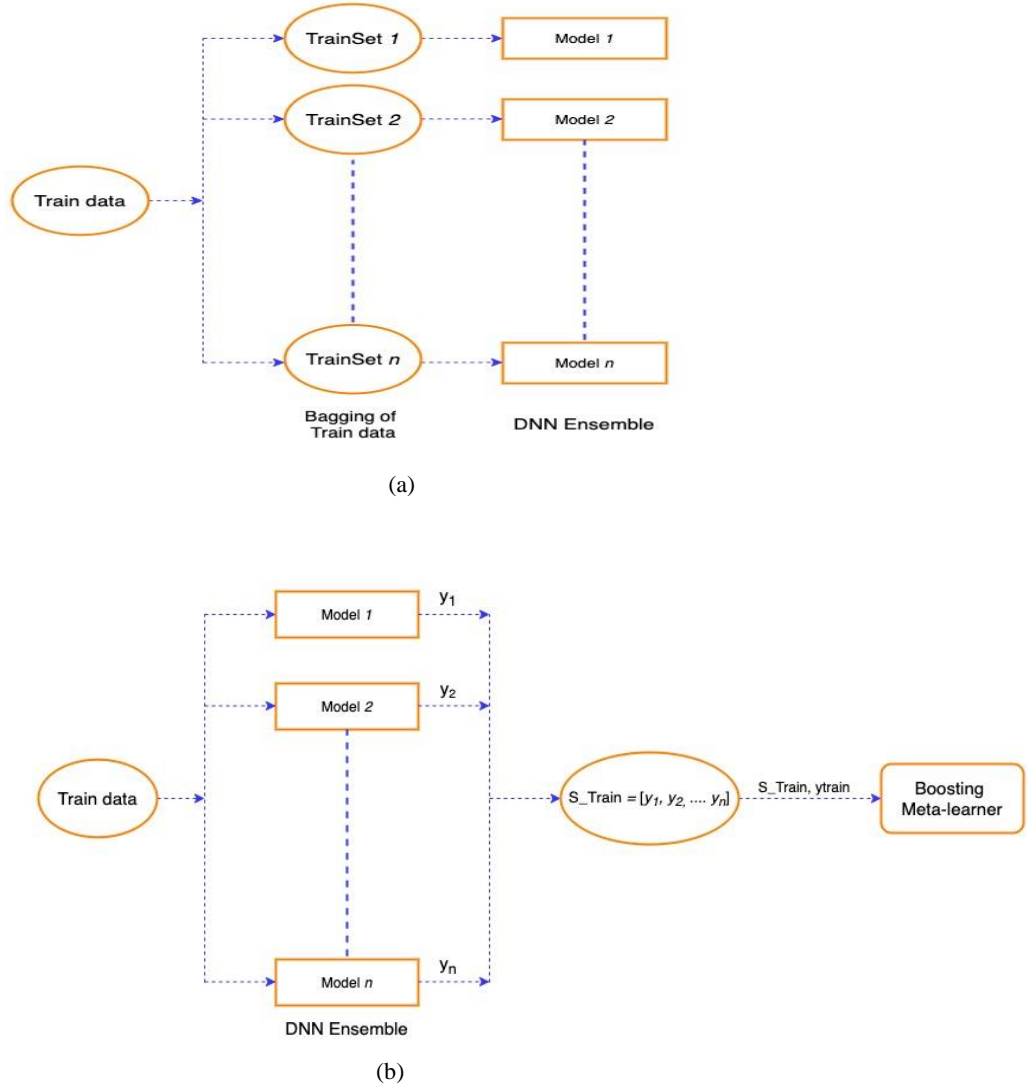

(a)

(b)

**Fig. 1.** DNN ensemble (a) training of individual DNNs (b) training the meta-learner for combining outputs of DNN ensemble

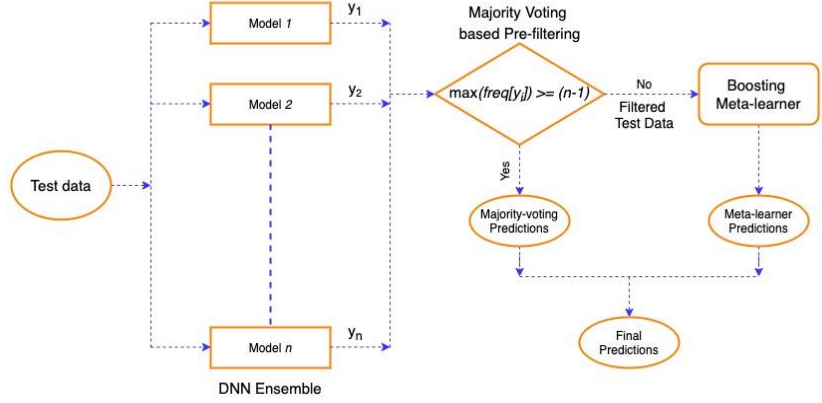

**Fig. 2.** Process flow for obtaining final prediction for a test sample.

## 4 Results

The software implementation was performed in Python 3.7 on an Intel dual-core processor. We conducted our experiments on five datasets namely, Human Activity Recognition (HAR) [20] and Gas sensor array drift, Isolet, Internet advertisements, and Spambase available in the UCI machine learning repository [21]. The HAR dataset is split into training and testing sets at the source. For the rest of the datasets, 80:20 split is used with the number of epochs set to 25. We choose an odd number of DNNs in our ensemble (=7). The optimal ensemble size was observed to be seven as demonstrated in the graphs in Fig. 3, drawn for the HAR dataset. The classification results are shown in Tables 1 (highest scores) and 2 (all fusion strategies). As observed from Table 2, all the fusion strategies of plurality voting, meta-learning and the proposed *pre-filtering by majority voting coupled with stacked meta-learner* gave accuracies that were better than the accuracy of the individual DNN and mean accuracy of all DNNs shown in Table 1. Our observations from Table 2 are: 1) Pre-filtering by majority voting increases the performance of meta-learning and overall gives a consistent performance 2) Meta-learning by itself is the second-best performer outperforming the proposed method in only one case out of five. 3) Plurality voting by itself does not perform as well. In the case of Spambase, the performance is unaffected by the choice of fusion strategy.

**Table 1.** Performance of DNN ensemble versus individual and mean DNN accuracies

| Name of dataset | Individual Accuracy | Mean accuracy | (Highest) Ensemble Accuracy (refer Table 2) |
|---|---|---|---|
| Human activity recognition (HAR) | 93.9% | 94.5% | 95.3% |
| Gas sensor array drift | 96.0% | 97.2% | 98.1% |
| Isolet | 94.5% | 94.0% | 96.2% |
| Internet advertisements | 97.2% | 97.5% | 98.1% |
| Spambase | 94.2% | 93.6% | 95.1% |

**Table 2.** Accuracy of decision fusion strategies (highest accuracy highlighted in bold)

| Name of dataset | *Plurality voting (maximum votes)* | *Meta-learning with XGBoost* | *Majority voting cum Meta-learning (proposed)* |
|---|---|---|---|
| Human activity recognition (HAR) | 95.0% | 95.0% | **95.2**% |
| Gas sensor array drift | 97.9% | **98.1**% | **98.1**% |
| Isolet | 96.0% | **96.2**% | 96.1% |
| Internet advertisements | 97.7% | 97.7% | **98.1**% |
| Spambase | **94.6**% | **94.6**% | **94.6**% |

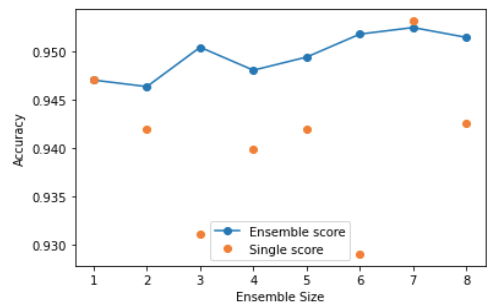

**Fig. 3.** Variation of accuracy of HAR dataset using proposed method with the size of ensemble (average accuracy of DNNs for each ensemble shown in orange dots).

## 5  Conclusion

In this paper we propose a homogeneous ensemble learning approach using DNNs. The training input was diversified for the seven DNNs by careful sampling. The fusion strategies of plurality voting, meta-learning and the proposed *pre-filtering by majority voting coupled with stacked meta-learner* result in better accuracies as compared to the individual accuracies of DNNs and their mean accuracies, for the five datasets. The proposed fusion method improved the results of meta-learning in most of the cases.